\def\paperID{0000}
\def\confName{CVPR}
\def\confYear{2025}

\def\paperTitle{FontUse: A Data-Centric Approach to Style- and Use-Case-Conditioned In-Image Typography}

\def\authorBlock{
    Xia Xin
    \qquad
     Yuki Endo
     \qquad
    Yoshihiro Kanamori \\
    University of Tsukuba \\
    {\tt\small cynthia160506@gmail.com \qquad \{endo, kanamori\}@cs.tsukuba.ac.jp}
}

\newif\ifreview 
\newif\ifarxiv \newcommand{\arxiv}{\arxivtrue}
\newif\ifcamera 
\newif\ifrebuttal 

\arxiv 

\pdfoutput=1
\documentclass[10pt,twocolumn,letterpaper]{article}
\ifreview \usepackage[review]{cvpr} \fi
\ifarxiv \usepackage[pagenumbers]{cvpr} \fi
\ifrebuttal \usepackage[rebuttal]{cvpr} \fi
\ifcamera \usepackage{cvpr} \fi

\usepackage{graphicx}	
\usepackage{amsmath}	
\usepackage{amssymb}	
\usepackage{booktabs}
\usepackage{times}
\usepackage{microtype}
\usepackage{epsfig}
\usepackage[table,xcdraw,dvipsnames]{xcolor}
\usepackage{caption}
\usepackage{float}
\usepackage{placeins}
\usepackage{color, colortbl}
\usepackage{stfloats}
\usepackage{enumitem}
\usepackage{tabularx}
\usepackage{xstring}
\usepackage{multirow}
\usepackage{xspace}
\usepackage{url}
\usepackage{subcaption}
\usepackage{xcolor}
\usepackage[hang,flushmargin]{footmisc}

\ifcamera \usepackage[accsupp]{axessibility} \fi

\usepackage{comment}
\usepackage{booktabs}
\usepackage{multirow}
\usepackage{graphicx}
\usepackage{xfrac}
\usepackage{placeins}
\usepackage[T1]{fontenc}
\usepackage{lmodern}
\usepackage{type1cm}
\usepackage{diagbox}
\usepackage{pifont}

\usepackage{tabularx}
\usepackage{booktabs}
\usepackage[labelfont=bf,format=plain,singlelinecheck=false]{caption}
\usepackage{multirow}

\usepackage{mathcomp}

\usepackage{color}

% ...

\ifarxiv  \fi

\newcommand{\R}[1]{{%
    \textbf{%
        \ifstrequal{#1}{1}{\textcolor{red}{R#1}}{%
        \ifstrequal{#1}{2}{\textcolor{blue}{R#1}}{%
        \ifstrequal{#1}{3}{\textcolor{magenta}{R#1}}{%
        \ifstrequal{#1}{4}{\textcolor{teal}{R#1}}{%
                           \textcolor{cyan}{R#1}%
        }}}}%
    }%
}}

\usepackage{xcolor}

\definecolor{mygreen}{RGB}{91,132,20}
\newcommand{\targetstring}[1]{\textcolor{mygreen}{#1}}

\definecolor{mybrown}{RGB}{155,80,13}
\newcommand{\fontstyle}[1]{\textcolor{mybrown}{#1}}

\definecolor{mycyan}{RGB}{40,125,206}
\newcommand{\usecase}[1]{\textcolor{mycyan}{#1}}

\definecolor{mypurple}{RGB}{153,102,204}
\newcommand{\effect}[1]{\textcolor{mypurple}{#1}}

\usepackage[T1]{fontenc}

\usepackage{booktabs,tabularx}
\usepackage{enumitem}
\usepackage{microtype}
\usepackage{wrapfig}  

\usepackage{xr-hyper}

\makeatletter
\newcommand*{\addFileDependency}[1]{
  \typeout{(#1)}
  \@addtofilelist{#1}
  \IfFileExists{#1}{}{\typeout{No file #1.}}
}

\makeatother
\newcommand*{\myexternaldocument}[1]{
    \externaldocument{#1}
    \addFileDependency{#1.tex}
    \addFileDependency{#1.aux}
}

\definecolor{cvprblue}{rgb}{0.21,0.49,0.74}
\usepackage[pagebackref,breaklinks,colorlinks,citecolor=cvprblue]{hyperref}
\usepackage[capitalize]{cleveref}
\crefname{section}{Sec.}{Secs.}
\crefname{table}{Table}{Tables}
\crefname{figure}{Fig.}{Figs.}

\ifarxiv \crefname{appendix}{App.}{Apps.}
\else \crefname{appendix}{Suppl.}{Suppls.} \fi

\unless\ifarxiv \myexternaldocument{_supplementary} \fi

\begin{document}
\title{\paperTitle}
\author{\authorBlock}
\twocolumn[{%
\renewcommand\twocolumn[1][]{#1}%
\maketitle
\includegraphics[width=1\linewidth]{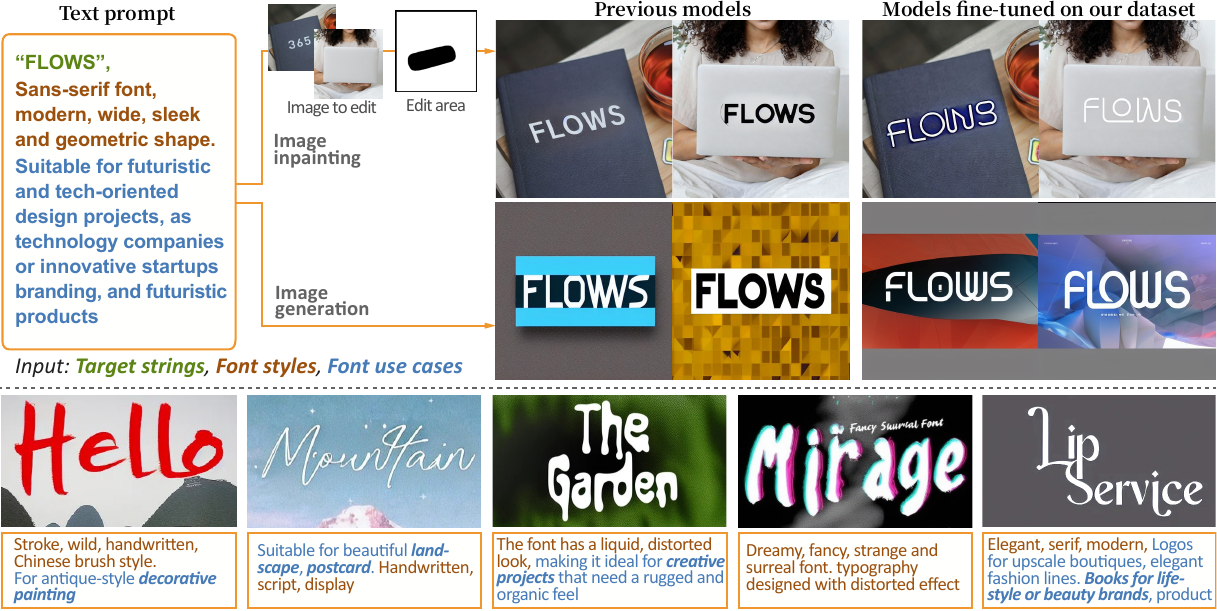}
\captionof{figure}{Fine-tuned models with our supervision (\emph{FontUse}) generate legible \targetstring{target strings} consistent with specified \fontstyle{font styles} and \usecase{use cases}. \textbf{Top}: Under identical prompts, fine-tuned models show improved results over baselines in both inpainting and full-image generation. \textbf{Bottom}: Additional examples illustrate diverse glyph forms and stylistic effects (backgrounds cropped; target strings omitted).}
\label{fig:teaser}
}]
\begin{abstract}
Recent text-to-image models can generate high-quality images from natural-language prompts, yet controlling typography remains challenging: requested typographic appearance is often ignored or only weakly followed. We address this limitation with a data-centric approach that trains image generation models using targeted supervision derived from a structured annotation pipeline specialized for typography. Our pipeline constructs a large-scale typography-focused dataset, \emph{FontUse}, consisting of about 70K images annotated with user-friendly prompts, text-region locations, and OCR-recognized strings. The annotations are automatically produced using segmentation models and multimodal large language models (MLLMs). The prompts explicitly combine \emph{font styles} (\eg, serif, script, elegant) and \emph{use cases} (\eg, wedding invitations, coffee-shop menus), enabling intuitive specification even for novice users. Fine-tuning existing generators with these annotations allows them to consistently interpret style and use-case conditions as textual prompts without architectural modification. For evaluation, we introduce a Long-CLIP-based metric that measures alignment between generated typography and requested attributes. Experiments across diverse prompts and layouts show that models trained with our pipeline produce text renderings more consistent with prompts than competitive baselines. The source code for our annotation pipeline is available at \url{https://github.com/xiaxinz/FontUSE}.
\end{abstract}
\section{Introduction}
Typography is central to visual communication and persuasion. Recent text-to-image models can synthesize complex scenes from natural-language prompts, including images containing text~\cite{dalle3,sd3,flux,FontDiffuser,Chen2023TextDiffuserDM,Tuo2023AnyTextMV}. However, despite advances in image quality and semantic composition, current systems still struggle to provide reliable and fine-grained control over rendered typography. In practice, prompts describing typographic intent, such as font appearance or intended usage context, are often ignored or only weakly reflected, forcing users to rely on trial-and-error prompt engineering.

A key reason for this limitation is insufficient supervision for typography-specific attributes. Existing training data typically provide only coarse semantic descriptions and lack structured annotations indicating text regions, textual content, or design-oriented attributes. As a result, models receive little guidance for typographic decisions and cannot consistently interpret user-specified design intentions.

To address this gap, we propose a data-centric framework for controllable in-image typography. Instead of modifying model architectures, we focus on constructing supervision signals tailored to typographic generation. Concretely, we introduce a scalable annotation pipeline that integrates segmentation, OCR, and multimodal large language models (MLLMs) to automatically produce structured labels aligned with typographic control. Using this pipeline, we assemble a large-scale collection of typography-focused images with annotations including text-region locations, recognized strings, and natural-language prompts describing typographic intent.

Our framework represents typographic intent along two complementary axes: \emph{font style}, describing visual characteristics (\eg, serif, handwritten, elegant), and \emph{use case}, specifying practical contexts in which the typography is appropriate (\eg, wedding invitations, product branding). While style provides abstract design guidance, specifying use cases makes user intent more concrete and accessible, particularly for non-experts. This dual-axis formulation enables intuitive prompt specification, improves consistency across generated assets, and reduces prompt iteration.

Importantly, our approach is architecture-independent for diffusion-based text generation models. The same supervision can be used to fine-tune existing text-capable image generators without architectural modification. The resulting models can render text that reflects both style and intended usage, either by inserting text into specified regions of existing images or by generating complete images containing typography (see Fig.~\ref{fig:teaser}).

To evaluate typography fidelity, we further introduce an automatic assessment framework tailored to this task. It includes a Long-CLIP~\cite{Zhang2024LongCLIPUT}-based metric that measures alignment between generated typography and textual specifications, as well as an MLLM-based pairwise preference evaluation that focuses on typographic attributes. Experiments across diverse prompts and layouts demonstrate that models trained with our supervision produce typography that better matches requested style and use-case conditions than competitive baselines.

In summary, our contributions are as follows:
\begin{itemize}
  \item A controllable typography generation framework enabling conditioning of in-image text on two complementary axes: font style and use case, 
  \item A scalable annotation pipeline integrating MLLM-based labeling, segmentation, and OCR, together with a large-scale typography dataset (\emph{FontUse}) of approximately 70,000 annotated font design images, and
  \item Typography-specific evaluation protocols, including a Long-CLIP-based alignment metric and an MLLM-based pairwise preference scheme, along with empirical validation of data-centric supervision.
\end{itemize}

\section{Related Work}
\label{sec:RelatedWork}
\noindent\textbf{Text rendering in image generation.}
Despite rapid progress in text-to-image diffusion models~\cite{Ho2020DenoisingDP,DhariwalN21,Rombach2021HighResolutionIS}, reliable and controllable rendering of in-image typography remains challenging. Early systems such as DALL-E~2~\cite{Ramesh2022HierarchicalTI} and Stable Diffusion~\cite{Rombach2021HighResolutionIS,SDXL} often produced illegible text. Later advances in large-scale variants~\cite{dalle3,sd3,flux} significantly improved visual fidelity and text legibility, demonstrating that modern generators can capture glyph structure at scale. However, explicit control over fine-grained typographic attributes remains limited, as prompts specifying style or intended usage are frequently interpreted loosely. This suggests that the remaining bottleneck lies less in model capacity than in supervision tailored to typography.

\noindent\textbf{Scene text generation.}
Several works improve text rendering by modifying the generative process. TextDiffuser~\cite{Chen2023TextDiffuserDM} and AnyText~\cite{Tuo2023AnyTextMV} introduce glyph- or layout-aware modules to guide character shapes and placement, while UDiffText~\cite{Zhao2023UDiffTextAU} enables region-constrained synthesis for inserting text into existing images. Although effective for lagibility and localization, these methods provide limited conditioning on typographic style or usage intent, restricting control over design-oriented attributes. 

\noindent\textbf{Typeface synthesis.}
A related line of research studies typeface synthesis, i.e., generating new fonts or glyph sets on clean canvases. Methods such as Attribute2Font~\cite{Attribute2Font}, DG-Font~\cite{DGfont}, and FontDiffuser~\cite{FontDiffuser} focus on controllable font design or few-shot glyph generation. These works address font creation as a design problem, whereas our setting is scene-level typography rendering conditioned on style and use-case attributes. Thus, the objectives and supervision requirements are fundamentally different.

\noindent\textbf{Typography supervision.}
Existing scene-text datasets such as SynthText~\cite{SynthText}, COCO-Text~\cite{COCOText}, ICDAR~\cite{ICDAR13}, and Total-Text~\cite{CK2019} are primarily designed for detection and recognition, providing geometric annotations and transcriptions but no style or usage information. The MyFonts-based dataset~\cite{Myfonts-dataset} includes style tags for retrieval, yet focuses on isolated glyph images without contextual or application-level descriptions. Consequently, current training corpora lack the supervision required for promptable typographic control. We address this limitation by constructing structured annotations that pair scene-like text images with style and use-case labels and editable regions.

\noindent\textbf{Automatic evaluation.}
Recent studies investigate large language models as evaluators for subjective tasks such as quality and preference assessment, with surveys examining reliability, bias, and protocol design~\cite{LLMasJudgeSurvey2024, Li2024LLMsasJudgesAC}. Empirical work shows that well-designed LLM-based evaluators can approximate human preferences~\cite{MTBench2023, Zhu2023JudgeLMFL}. In multimodal settings, MLLM-based judges support scoring and pairwise comparison, with evidence that pairwise evaluation achieves stronger agreement with human judgments~\cite{MLLMasJudge2024, MTBench2023}. Following these findings, we adopt MLLM-based pairwise evaluation to assess typography faithfulness alongside human studies.
\section{Method}
\label{sec:OurMethod}

\begin{figure*}[t]
\centering
\includegraphics[width=1.\textwidth]{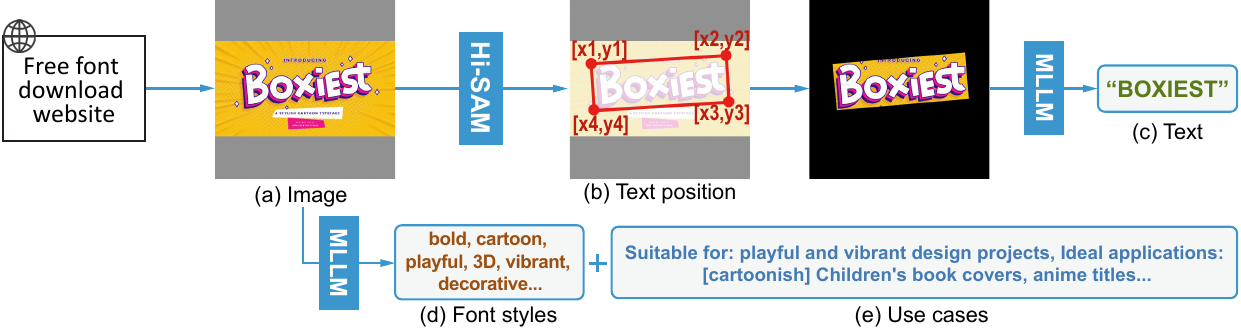}
\caption{
Dataset construction pipeline. (a) Typography-focused images are collected from public design resources. (b) Hi-SAM~\cite{Ye2024HiSAMMS} detects text regions and outputs bounding boxes. (c)-(e) An MLLM performs text recognition, font-style annotation, and use-case annotation.}
\label{fig:dataset}
\end{figure*}

\noindent\textbf{Problem setting and approach.} 
Our goal is controllable typography generation for both in-image editing and full-image synthesis, where the model must render text that is legible and consistent with requested style and usage conditions. We address this challenge from a data-centric perspective: instead of modifying architectures, we construct structured supervision tailored to typographic control and use it to fine-tune existing generators. This design does not depend on architecture-specific modifications and transfers across different diffusion backbones. Specifically, we adapt AnyText~\cite{Tuo2023AnyTextMV} with our supervision and show that the same training recipe generalizes to TextDiffuser-2~\cite{chen2023textdiffuser-2} and Stable Diffusion~3~\cite{sd3}, consistently improving alignment and legibility.

\begin{figure*}[t]
\centering
\includegraphics[width=0.8\linewidth]{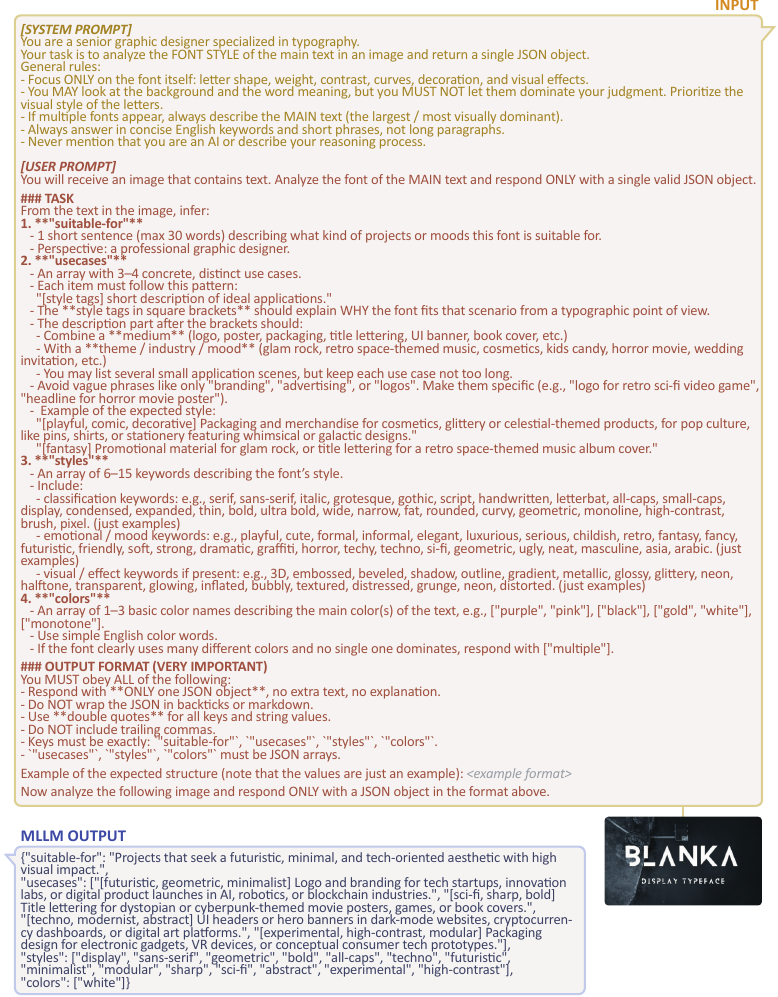}
\caption{
Example of the input prompt and an MLLM responses. See Appendix~\ref{app:appendix-gpt} for details of \textlangle example format\textrangle.
}
\label{fig:gpt}
\end{figure*}

\subsection{Dataset Construction}
\label{dataset-gen}
Fig.~\ref{fig:dataset} shows the data construction pipeline. We collect 70,000 typography images from a publicly available font-design website~\cite{1001fonts} and automatically attach structured annotations, including text-region boxes, OCR strings, and style/use-case descriptions.

\noindent\textbf{Text region localization and recognition.}
We adopt a two-stage pipeline: word-level detection followed by recognition. Hi-SAM~\cite{Ye2024HiSAMMS} produces word-level masks, which are converted into normalized bounding boxes for region extraction. Each cropped word region is independently recognized using an MLLM, enabling robust recognition of stylized or visually complex text. The prompt design and its rationale are described in Sec.~\ref{subsec:prompt}.

\noindent\textbf{MLLM-based annotation.}
We use an MLLM to automatically generate style and use-case descriptions for each image using a structured prompt template, as manual annotation would require substantial effort and domain expertise. The generated outputs are paired with images to form supervision for fine-tuning. To improve reliability, the template enforces a fixed JSON schema and constrained outputs, reducing ambiguity and noise in the annotations. 

\noindent\textbf{Annotation schema.}
Each annotation is represented as a JSON object with four fields: \texttt{suitable-for}, \texttt{usecases}, \texttt{styles}, and \texttt{colors}. These fields capture semantic intent, application context, stylistic attributes, and dominant text colors. By separating these factors, the schema encourages disentangled typographic control. During training, all fields are concatenated as conditioning text, while at inference time any subset can be used for flexible control.

\subsection{Prompt Design and Rationale}
\label{subsec:prompt}

We design prompts to generate structured typography annotations suitable for large-scale supervision. The design enforces three constraints: (i) font-focused descriptions, (ii) primary-text selection, and (iii) structured outputs. Font-focused descriptions restrict annotations to intrinsic typographic attributes (\eg, stroke shape, weight, curvature, ornamentation, and visual effects). Primary-text selection limits annotation to the dominant text region. Structured outputs require a single JSON object with fixed keys and schema validation. These constraints are realized through separate system and user prompts: the system prompt defines invariant behavioral rules, while the user prompt specifies the task and output schema (Fig.~\ref{fig:gpt}). Together, these constraints reduce annotation ambiguity and improve consistency across generated supervision.

\begin{figure}[t]
\centering
\includegraphics[width=1\linewidth]{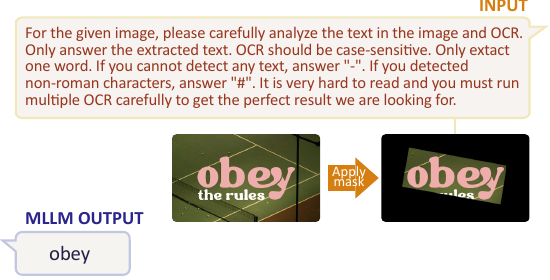}
\caption{Prompt used for text recognition. The MLLM is instructed to output only the extracted word (case-sensitive), returning ``-'' if no text is detected and ``\#'' for non-Roman characters.}
\label{fig:gpt-ocr}
\end{figure}

\noindent\textbf{System prompt.}
We use a system prompt to enforce fixed annotation constraints, ensuring consistent behavior across all annotation queries. It specifies (1) a typographic-expert role, (2) font-only evaluation scope, and (3) restricted output format, designed to minimize irrelevant variation in generated annotations. Concretely, the prompt instructs the model to act as a typography specialist, focus only on glyph attributes, and produce concise keyword outputs without reasoning text.

\noindent\textbf{User prompt.}
The user prompt provides task instructions and enforces the predefined JSON format, complementing the system prompt’s global constraints. By requiring structured outputs, it ensures consistent and machine-readable annotations for supervision.

\noindent\textbf{OCR prompt.}
We design a dedicated OCR prompt (Fig.~\ref{fig:gpt-ocr}) for word-level recognition, separate from the annotation prompt. The prompt constrains the model to output only a single extracted word without explanation and enforces case-sensitive transcription. 
To facilitate reliable filtering, failure cases are standardized: the model must return ``-'' for recognition failure and ``\#'' for non-Roman text. These predefined tokens enable consistent post-processing. This design improves robustness to visually ambiguous or highly stylized glyphs, where conventional OCR systems tend to degrade (Sec.~\ref{sec:ocr-eval}).

\subsection{Dataset Composition and Statistics}
We summarize dataset composition using statistics derived from JSON annotations. For each record, we parse the \texttt{styles} and \texttt{usecases} fields and split use-case descriptions into short noun phrases. Both sources undergo light normalization (\eg, case folding and synonym mapping). We then encode unique phrases with a CLIP~\cite{Radford2021LearningTV} text encoder and perform a thresholded union-find clustering, where each component defines a family labeled by its medoid phrase (threshold = 0.9). We report family frequencies to analyze head-tail behavior.

\begin{figure*}[t]
  \centering
  \includegraphics[width=1\linewidth]{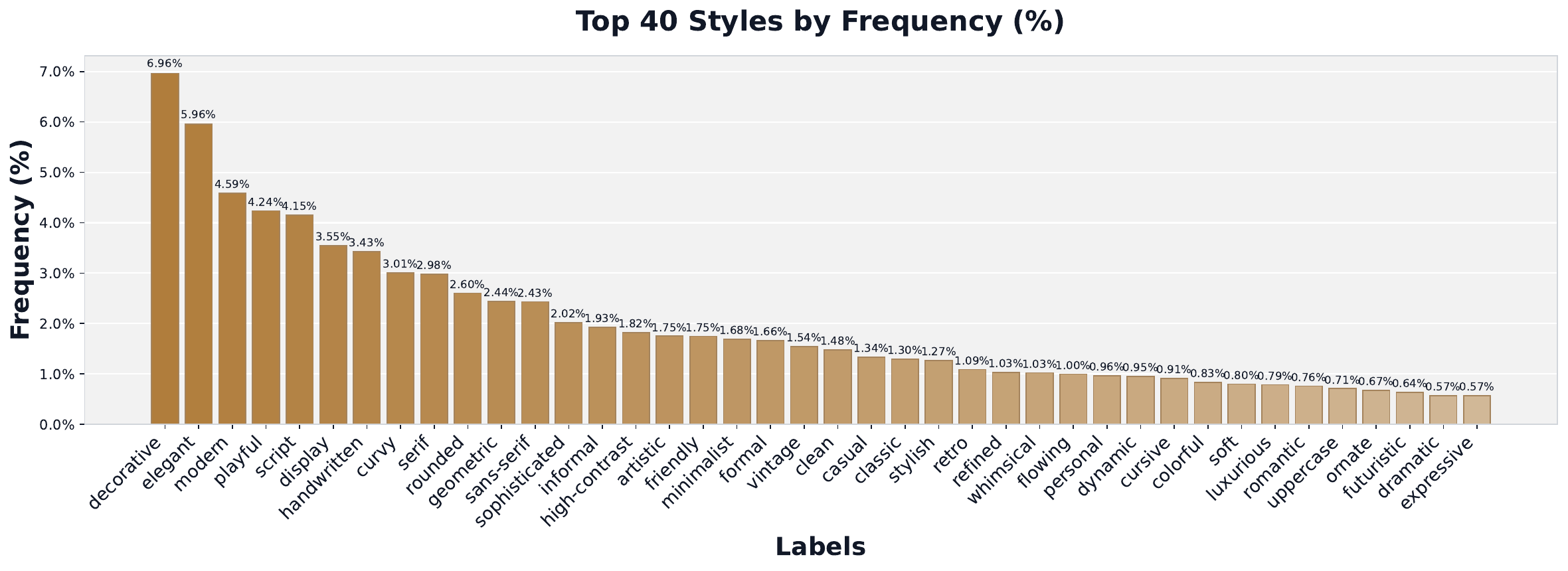}
  \includegraphics[width=1\linewidth]{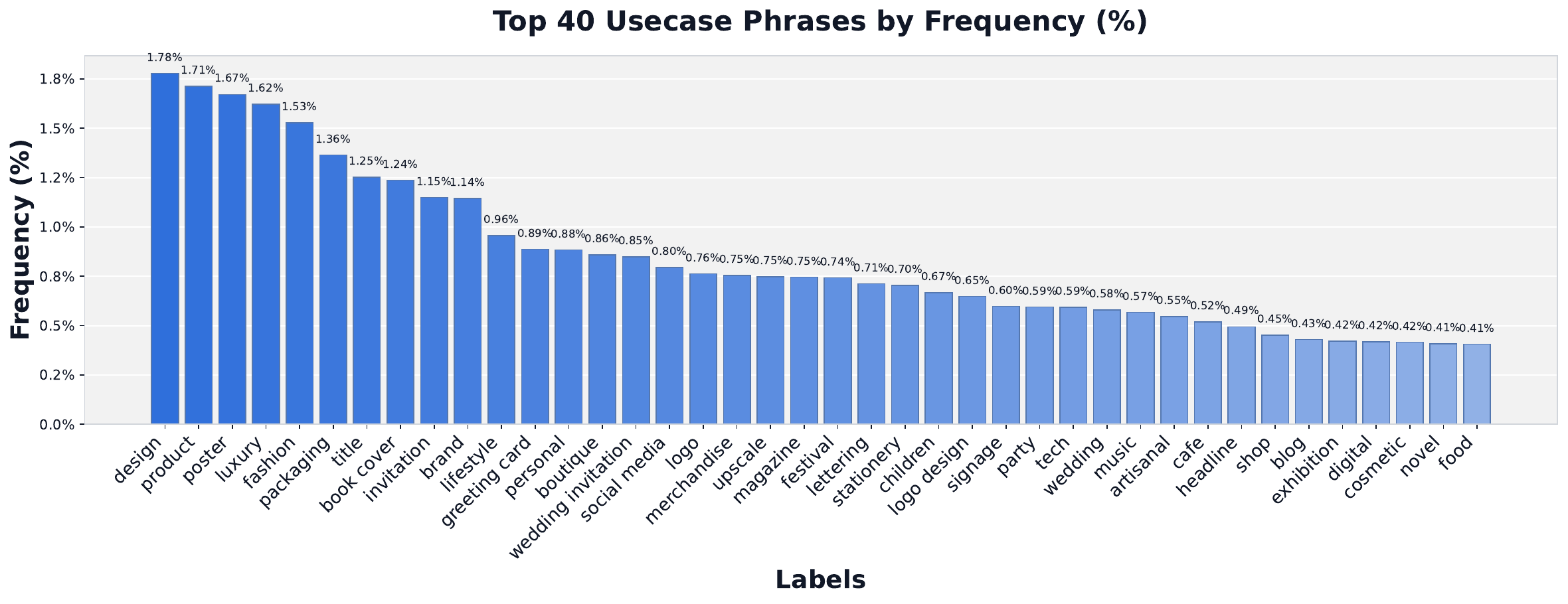}
  \caption{Top-40 style and use-case families after consolidation, illustrating their frequency distributions.}
  \label{fig:stats-top}
\end{figure*}

\begin{figure}[t]
  \centering
  \includegraphics[width=1\linewidth]{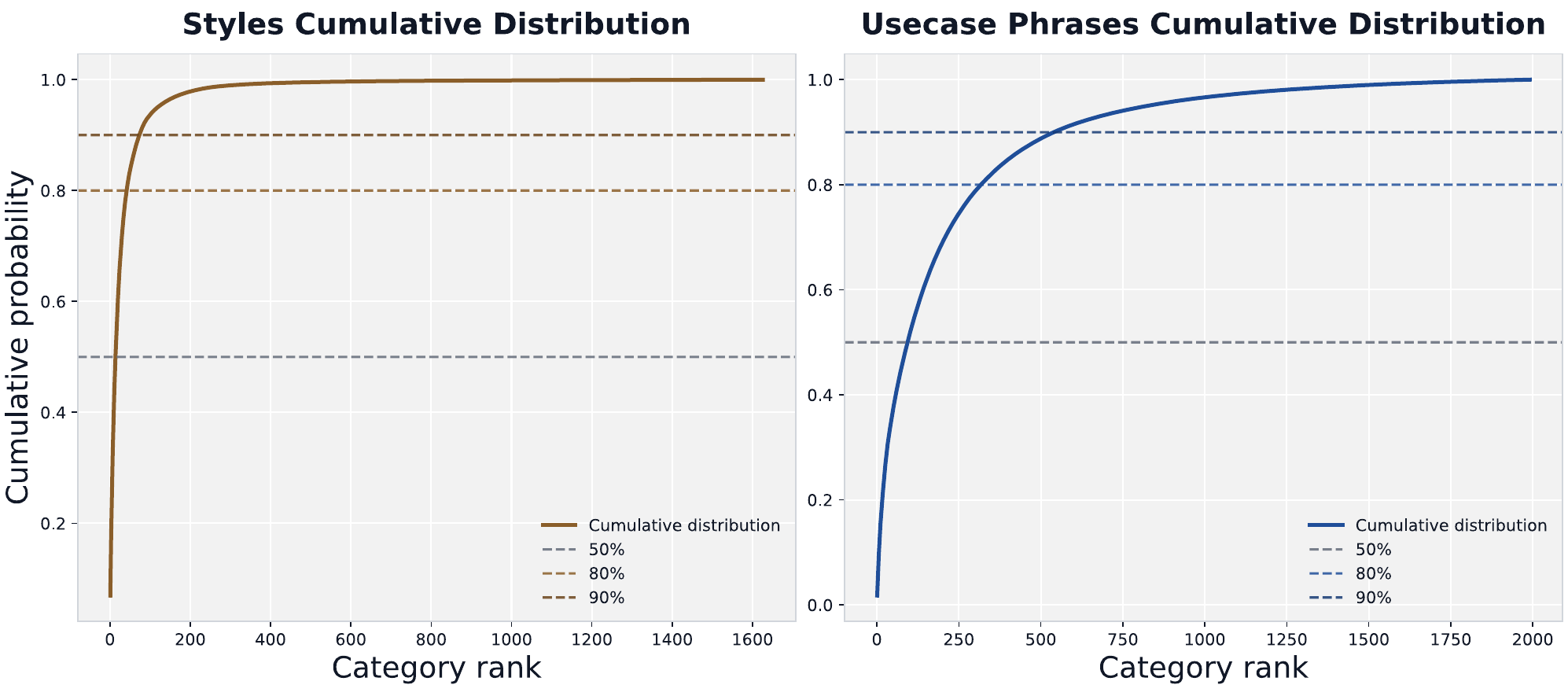}
  \caption{
  Cumulative distributions of use-case and style families. Dashed lines indicate 50\%, 80\%, and 90\% cumulative mass.
  }
  \label{fig:stats-cdf}
\end{figure}

Fig.~\ref{fig:stats-top} shows the most frequent categories, while Fig.~\ref{fig:stats-cdf} summarizes overall coverage. Together, these statistics indicate that common typographic attributes are well represented while long-tail categories remain broadly covered. The CLIP-based consolidation yields compact, interpretable families suitable for supervision and evaluation.

\section{Experiments}\label{sec:eval}

\subsection{Implementation Details}
We used GPT-4o~\cite{openAI} for annotation generation and GPT-4o-mini~\cite{openAI} for OCR inference. We fine-tuned models using PyTorch on an NVIDIA RTX A6000 with 40{,}000 image-text prompt pairs from our dataset (the remaining 30{,}000 are used as described in Sec.~\ref{sec:LongCLIP}). All models were trained under the same settings: AdamW~\cite{DBLP:conf/iclr/LoshchilovH19}, learning rate $2\times10^{-5}$, batch size 10, and 60 epochs. Training took about 10 days, and inference required 2--5 seconds per $512\times512$ image.

\subsection{Evaluation Metrics}
\label{sec:LongCLIP}
\noindent\textbf{Fine-tuned Long-CLIP.}
To evaluate the alignment between generated text images and the input font styles/use cases, we employ Long-CLIP~\cite{Zhang2024LongCLIPUT}, which supports longer text inputs than CLIP~\cite{Radford2021LearningTV}. As discussed later, Long-CLIP provides a clearer measure of similarity between text images and style/use-case descriptions than CLIP. However, because Long-CLIP is trained on generic image data and not specialized for font features, we further fine-tune it for typographic evaluation.
Specifically, we fine-tuned Long-CLIP with 30{,}000 image-prompt pairs from our dataset that were \emph{not} used to train the generation models. To exclude background influences during evaluation, we extract only the text regions from images and perform contrastive learning between the cropped text images and their corresponding prompts using the pretrained Long-CLIP. The benefit of fine-tuning is validated in Sec.~\ref{subsec:metric-validation}.

\noindent\textbf{MLLM-based pairwise preference.}
To complement embedding scores with preference judgments focused on typography, we employ two multimodal LLMs, GPT-4o~\cite{openAI} and Gemini 2.5 Pro~\cite{gemini-ref}. For each backbone, we form two random pairs per prompt, each pair containing one image from the base model and one from its fine-tuned counterpart (order randomized). The instruction constrains the judge to evaluate \emph{only} the typography and to ignore backgrounds. The full instruction is described in Appendix~\ref{app:appendix-mllm-eval}. The validity of the MLLM-based evaluations is further examined in Sec.~\ref{subsec:metric-validation}, where we compare them against human judgments.

\noindent\textbf{Legibility. }
We evaluate legibility using two human assessors who independently transcribe the generated text. For each image, we compute the character error rate (CER) between each transcription and the corresponding target string as the Levenshtein edit distance normalized by the length of the target string. The final legibility score is obtained by averaging CER across assessors and images.


\subsection{Comparison with Baselines}

\noindent\textbf{Evaluation protocol.}
We evaluate whether fine-tuning with our dataset improves text-image alignment and preference. We compare AnyText~\cite{Tuo2023AnyTextMV}, TextDiffuser-2 (TD-2)~\cite{chen2023textdiffuser-2}, and Stable Diffusion~3 (SD-3)~\cite{sd3} with their fine-tuned counterparts (\emph{Fine-tuned-AnyText}, \emph{Fine-tuned-TD-2}, \emph{Fine-tuned-SD-3}). For each model, we generate 10 images for each of 30 prompts (see Appendix~\ref{app:appendix-prompts} for the prompt list) and crop the text regions to isolate generated typography. To quantify text-image alignment, we report three similarity metrics: (i) CLIP, (ii) the original Long-CLIP, and (iii) our fine-tuned Long-CLIP. Each score is computed for every image-prompt pair and then averaged across all samples.
For preference evaluation, we randomly sampled 60 baseline-fine-tuned pairs (2 images per prompt across 30 prompts) and query GPT-4o and Gemini~2.5 Pro to select the preferred image. Legibility is evaluated on 120 randomly selected images per model (4 images per prompt across the same 30 prompts).

\noindent\textbf{Quantitative comparison.}
Tab.~\ref{tab:overall-longclip} summarizes alignment scores, MLLM preference, and legibility. Using our fine-tuned Long-CLIP evaluator, the fine-tuned models achieve the best alignment within each backbone, with the largest margins over the baselines. This trend is also observed with the general-purpose CLIP and the original Long-CLIP evaluators, although the separations are generally smaller than with our specialized evaluator. Both MLLM judges substantially favor the fine-tuned outputs, indicating that the improved embedding alignment is reflected in preference judgments focused on typography. Despite stronger stylistic rendering after fine-tuning, legibility does not degrade and generally improves.

\begin{table*}[t]
\centering
\caption{Comparison of CLIP, original Long-CLIP and fine-tuned Long-CLIP scores (linearly normalized to $[0,1]$), MLLM preference counts, and legibility (measured by the average character error rate, CER). Boldface indicates the best result within each backbone. 
FT denotes ``fine-tuned.''}
\label{tab:overall-longclip}
\begin{tabular}{lcc|cc|cc}
\toprule
& \multicolumn{2}{c}{AnyText} & \multicolumn{2}{c}{TD-2} & \multicolumn{2}{c}{SD-3} \\
\cmidrule(lr){2-3}\cmidrule(lr){4-5}\cmidrule(lr){6-7}
& base & FT & base & FT & base & FT \\
\midrule
CLIP $\uparrow$ & \textbf{0.3644} & 0.3241 & 0.4364 & \textbf{0.4550} & 0.4189 & \textbf{0.4355} \\
Long-CLIP $\uparrow$ & 0.3946 & \textbf{0.4464} & 0.4765 & \textbf{0.5170} & 0.4696 & \textbf{0.5618} \\
Fine-tuned Long-CLIP $\uparrow$ & 0.5858 & \textbf{0.6368} & 0.6097 & \textbf{0.7414} & 0.5068 & \textbf{0.6297} \\
\midrule
GPT selections $\uparrow$ & 12/60 & \textbf{48/60} & 9/60 & \textbf{51/60} & 16/60 & \textbf{44/60} \\
Gemini selections $\uparrow$ & 14/60 & \textbf{46/60} & 11/60 & \textbf{49/60} & 16/60 & \textbf{44/60} \\
\midrule
Legibility (Average CER)$\downarrow$ & 9.68\% & \textbf{5.91\%} & 2.72\% & \textbf{2.31\%} & 19.07\% & \textbf{8.29\%} \\
\bottomrule
\end{tabular}
\end{table*}

\begin{figure*}[t]
\centering
\includegraphics[width=\textwidth]{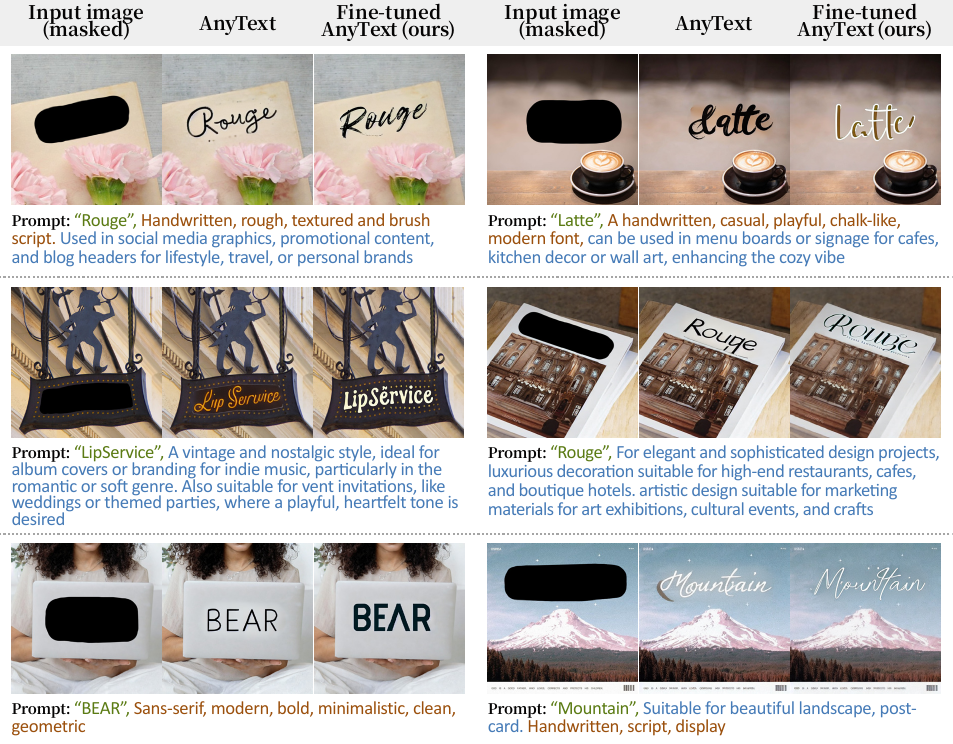}
\caption{Qualitative comparison with AnyText~\cite{Tuo2023AnyTextMV} and FontUse-tuned model (AnyText fine-tuned using our dataset). In the text prompts, the target string to be rendered as a text image is shown in {\targetstring{green}}, the font style in {\fontstyle{brown}}, and the font use case in {\usecase{blue}}.}
\label{fig:expriment-ex}
\end{figure*}

\begin{figure*}[t]
\centering
\includegraphics[width=\textwidth]{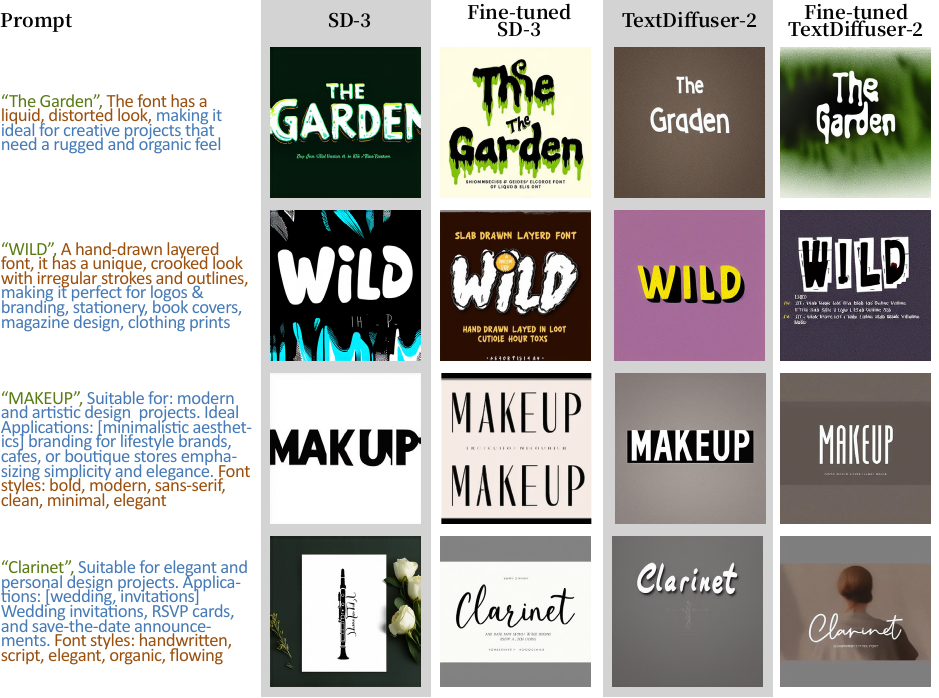}
\caption{Qualitative comparison with Stable Diffusion 3 (SD-3)~\cite{sd3}, TextDiffuser-2~\cite{chen2023textdiffuser-2}, and FontUse-tuned models (SD-3 and TextDiffuser-2 fine-tuned using our dataset).}
\label{fig:expriment-other}
\end{figure*}

\begin{figure*}[t]
\centering
\includegraphics[width=\textwidth]{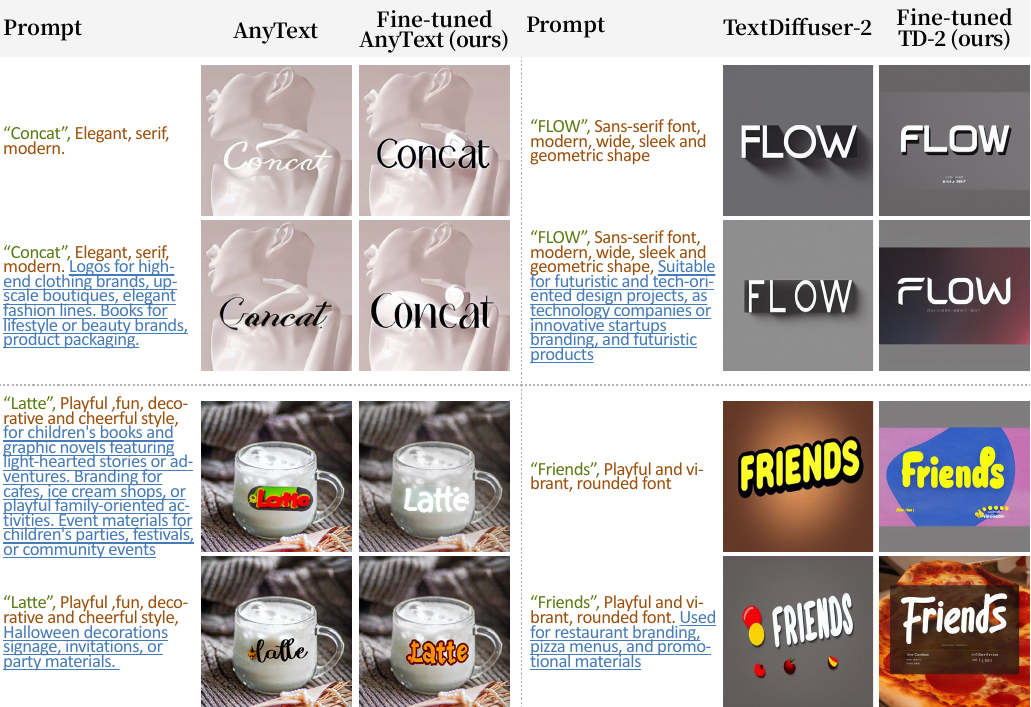}
\caption{Comparison of outputs across different use-case conditions. Underlined text indicates prompt differences. Input image and edit region omitted.}
\label{fig:expriment-usecase}
\end{figure*}

\begin{figure*}[t]
\centering
\includegraphics[width=\textwidth]{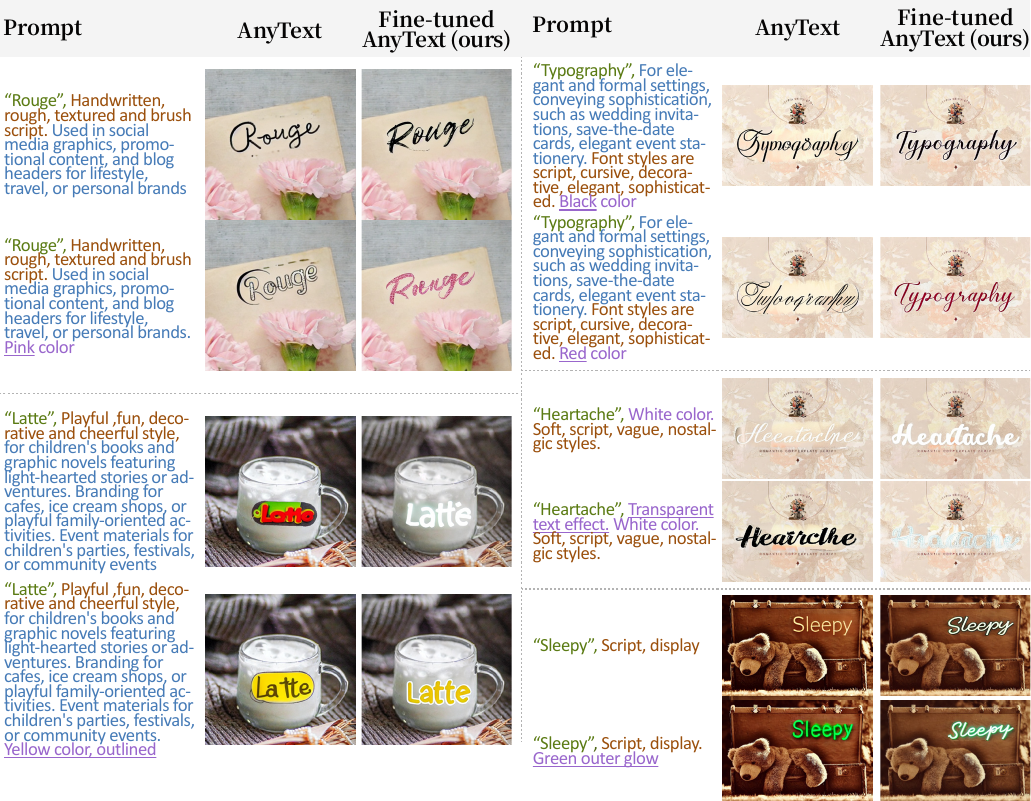}
\caption{Comparison of controllability under color and effect specifications (shown in {\effect{purple}}). Input image and edit region omitted.}
\label{fig:expriment-color}
\end{figure*}

\noindent\textbf{Qualitative comparison.}
The results are shown in Figs.~\ref{fig:expriment-ex} and \ref{fig:expriment-other}. Compared with the baselines, FontUse-tuned models better preserve legibility while more faithfully reflecting the specified font attributes.

To assess the effect of use-case conditioning, we compared outputs generated with identical font-style prompts while varying or removing only the use-case specification. Results in Fig.~\ref{fig:expriment-usecase} show that baseline models exhibit little visual difference regardless of use-case input, whereas our method adapts glyph characteristics to better match the specified context.

Our method also enables controllable rendering of color and physical text effects. As shown in Fig.~\ref{fig:expriment-color}, attributes such as embossed and glowing are reproduced as visible texture variations, which baseline models often fail to reflect.

\subsection{Reliability Analysis of Evaluation Metrics}
\label{subsec:metric-validation}
\noindent\textbf{Fine-tuned Long-CLIP.}
We validate whether the fine-tuned Long-CLIP~\cite{Zhang2024LongCLIPUT} can serve as a reliable metric for style/use-case alignment. A reliable metric should assign higher similarity to the correct paired use-case text than to non-paired texts, and lower similarity to an unrelated text. To test this, we randomly sampled 300 pairs from 40{,}000 image-text pairs that were \emph{not} used to train Long-CLIP. Across these samples, we computed the following average similarity scores using CLIP~\cite{Radford2021LearningTV}, Long-CLIP before fine-tuning, and Long-CLIP after fine-tuning:
\begin{enumerate}[label=(\alph*)]
\item \textbf{Similarity to the corresponding use-case text:} the average similarity between each image and its paired use-case text.
\item \textbf{Similarity to non-corresponding use-case texts:} the average similarity between each image and the texts corresponding to the other 299 images.
\item \textbf{Similarity to an unrelated text:} the average similarity between each image and the unrelated text ``A photo of a cat.''
\end{enumerate}

\begin{table*}[t]
\centering
\caption{Validation of the alignment metrics. We report the average similarity scores, computed by each metric between images and texts under three conditions: (a) corresponding use-case text, (b) non-corresponding use-case text, and (c) unrelated text.}
\label{tab:clips}
\begin{tabular}{lc|c|c}
\toprule
                   & CLIP & Long-CLIP & Fine-tuned Long-CLIP \\
\midrule
(a) corresponding & 22.13 & 65.31 & 72.76 \\
(b) non-corresponding & 21.71 & 61.40 & 62.72 \\
(c) unrelated text & 19.88 & 38.98 & 33.56 \\
\midrule
Gain (a vs. b) $\uparrow$ & +1.9 \%  & +6.4 \%  & \textbf{+16.0 \%}  \\
Gain (a vs. c) $\uparrow$ & +11.3 \% & +67.5 \% & \textbf{+116.8 \%} \\
\bottomrule
\end{tabular}
\end{table*}

As shown in Tab.~\ref{tab:clips}, the fine-tuned Long-CLIP yields higher similarity to the corresponding use-case text and lower similarity to the unrelated text compared with the pre-fine-tuning model. While similarity to non-corresponding use-case texts increased slightly, the change is small. In contrast, CLIP shows little separation among the three conditions, suggesting that it is less suitable for evaluating font use cases.

\noindent\textbf{MLLM-based pairwise preference.}
We conducted a user study to assess whether MLLM preferences track human judgments. Ten participants (five with graphic design experience) were shown 16 image pairs (backgrounds cropped to avoid bias) together with a style/use-case description and asked to choose which typography better matched the description; majority vote per pair formed the human reference. We then prompted GPT-4o and Gemini 2.5 Pro with the \emph{same} 16 pairs and instruction, recording the final choice. The two MLLMs agreed with each other on all 16 pairs and matched the human majority on 15/16 pairs (disagreeing on one borderline case), suggesting that MLLM-based preference evaluation can serve as a reasonable proxy for human judgments in our setting.

\subsection{Evaluation of the OCR Pipeline}
\label{sec:ocr-eval}
To validate the effectiveness of our text recognition process, we conducted a quantitative comparison with several OCR baselines~\cite{TesseractOCR,easyocr,PPOCRv3,olmOCR}.
We sampled 100 images from the dataset and manually annotated the ground-truth (GT) text for each image. We then computed the character error rate (CER) between each OCR output and the corresponding GT string using the same metric defined in Sec.~\ref{sec:LongCLIP}.

Tab.~\ref{tab:ocr-eval-res} shows the mean CER over the 100 images. Our method substantially reduces the error rate compared to traditional OCR engines, indicating that it is more robust to highly stylized fonts.

\begin{table}[t]
\centering
\caption{OCR performance measured by average CER.}
\label{tab:ocr-eval-res}
\begin{tabular}{@{}lc@{}}
\toprule
Method & Average CER$\downarrow$ \\
\midrule
Tesseract OCR~\cite{TesseractOCR} & 71.88\% \\
EasyOCR~\cite{easyocr} & 36.23\% \\
PaddleOCR(v3)~\cite{Li2022PPOCRv3MA} & 12.44\% \\
olmOCR-7B~\cite{olmOCR} & 3.78\% \\
Ours      & \textbf{2.18\%}  \\
\bottomrule
\end{tabular}
\end{table}

\section{Conclusions}
\label{sec:conclusions}
We introduced a data centric framework for controllable typography generation based on structured supervision. We argue that the main bottleneck in typography control is not model capacity, but the absence of supervision tailored to typographic intent. By modeling font style and use case as complementary conditioning axes and constructing scalable annotations, we enable consistent control over design oriented properties. Experiments across multiple diffusion backbones show that structured typographic supervision substantially improves alignment with user specified style and use cases while preserving legibility. 

\begin{figure}[t]
  \centering
  \includegraphics[width=\linewidth]{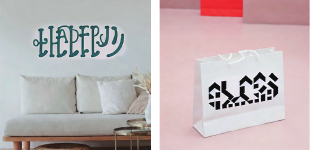}
  \caption{Our failure cases. There is a trade-off between stylistic complexity and legibility.}
  \label{fig:failure_cases}
\end{figure}

\noindent\textbf{Limitations and future work.}
Our system currently focuses on English typography, and multilingual extension remains future work. We also observe a trade-off between stylistic complexity and legibility: highly abstract or decorative styles may occasionally impair legibility (Fig.~\ref{fig:failure_cases}). Similar behavior is present in the base backbones, indicating that glyph structure preservation remains an open challenge.

{\small
\bibliographystyle{ieeenat_fullname}
\bibliography{11_references}
}

\ifarxiv \clearpage \appendix \appendix
\section{Details of Input Prompts}
\label{app:appendix-gpt}
To ensure accurate and machine-readable annotations, the input prompt includes an explicit example demonstrating the target output format. The full \textlangle example format\textrangle, omitted for brevity from the prompt illustration in Fig.~\ref{fig:gpt} of the main paper, is provided in Fig.~\ref{fig:gpt-prompt-example-details}.

\section{Details of Prompts for MLLM Evaluation}
\label{app:appendix-mllm-eval}
To obtain reliable typography-focused preference judgments from MLLMs, in the prompt we employ a set of instructions that (i) restrict attention to the text region, (ii) ignore backgrounds and design elements, (iii) require a single A/B choice with a brief justification, and (iv) enforce a fixed JSON output schema. The full prompt is shown in Fig.~\ref{fig:mllm-eval-prompt}.

\section{Details of Evaluation Prompts}
\label{app:appendix-prompts}
For quantitative evaluation, we curated a set of 30 prompts spanning diverse context and formulations. This set includes short and long phrasings, prompts specifying only \emph{styles}, only \emph{use cases}, and prompts specifying both. Tabs.~\ref{tab:prompts-1} and \ref{tab:prompts-2} list the complete set used in our experiments.

\begin{figure}[htbp]
\centering
\includegraphics[width=1\linewidth]{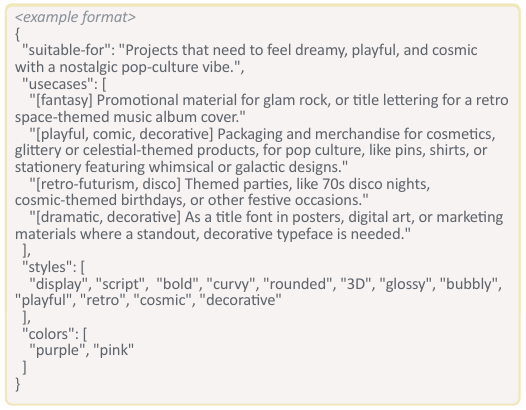}
\caption{Details of the required \textlangle example format\textrangle\ to enforce strict compliance with the target format.}
\label{fig:gpt-prompt-example-details}
\end{figure}

\begin{figure}[htbp]
\centering
\includegraphics[width=1\linewidth]{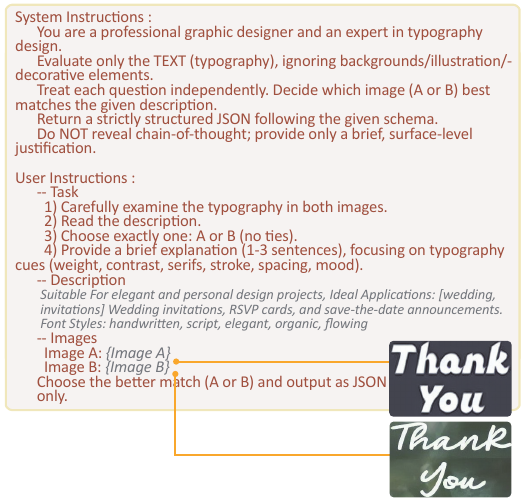}
\caption{Prompt for the MLLM-based evaluation. The detailed instructions are crafted to produce reliable typography-focused preference judgments.}
\label{fig:mllm-eval-prompt}
\end{figure}

\begin{table*}[htbp]
\centering
\caption{Evaluation prompts (ID: 1-15). \fontstyle{Font styles} are indicated in \fontstyle{brown} and \usecase{use cases} in \usecase{blue}.}
\label{tab:prompts-1}
\small
\setlength{\tabcolsep}{6pt}
\begin{tabularx}{\textwidth}{@{}c >{\raggedright\arraybackslash}X@{}}
\toprule
\textbf{ID} & \textbf{Prompt} \\
\midrule
1  & \usecase{Suitable for romantic and playful design projects, fun and flirty which fit brandings of dating services or Valentine's Day promotions.} \fontstyle{Cute, pretty, vague, nostalgic} \\
2  & \fontstyle{The font has a liquid, distorted look,} \usecase{making it ideal for creative projects that need a rugged and organic feel} \\
3  & \fontstyle{A hand-drawn layered font, it has a unique, crooked look with irregular strokes and outlines, } \usecase{making it perfect for logos \& branding, stationery, book covers, magazine design, clothing prints} \\
4  & \usecase{For advertising campaigns focused on youthful, casual consumers, targeting parties or social gatherings} \\
5  & \fontstyle{Sans-serif, modern, bold, minimalistic, clean, geometric} \\
6  & \usecase{Suitable for: modern and artistic design projects. Ideal applications: [minimalistic aesthetics] branding for lifestyle brands, cafes, or boutique stores emphasizing simplicity and elegance. } \fontstyle{Font styles: bold, modern, sans-serif, clean, minimal, elegant} \\
7  & \usecase{Suitable for elegant and personal design projects. Ideal applications: [wedding, invitations] wedding invitations, RSVP cards, and save-the-date announcements. } \fontstyle{Font styles: handwritten, script, elegant, organic, flowing} \\
8  & \usecase{For decorative art, like wall art for home or office spaces aimed at encouraging positivity and ambition} \\
9  & \fontstyle{Playful, fun, decorative and cheerful style.} \usecase{For children's books, comics, and graphic novels featuring light-hearted stories or adventures. Branding for cafes, ice cream shops, or playful family-oriented activities. Event materials for children's parties, festivals, or community events} \\
10 & \usecase{For elegant and formal settings, conveying sophistication, such as wedding invitations, save-the-date cards, elegant event stationery.} \fontstyle{Font styles are script, cursive, decorative, elegant, sophisticated} \\
11 & \fontstyle{A sans-serif, bold, modern, clean, minimalistic font.} \usecase{Suitable for modern and clean design projects, like Invitations for formal events or business meetings} \\
12 & \fontstyle{A vintage and nostalgic style,} \usecase{ideal for album covers or branding for indie music, particularly in the romantic or soft genre. Also suitable for event invitations, like weddings or themed parties, where a playful, heartfelt tone is desired} \\
13 & \usecase{Suitable for beautiful landscape.} \fontstyle{Postcard. handwritten, script, display} \\
14 & \usecase{For elegant and sophisticated design projects, luxurious decoration suitable for high-end restaurants, cafes, and boutique hotels. Artistic design suitable for marketing materials for art exhibitions, cultural events, and crafts} \\
15 & \usecase{For casual, welcoming, and creative coffee shop environments.} \fontstyle{A handwritten, casual, playful, chalk-like, modern font,} \usecase{can be used in menu boards or signage for coffee shops and cafes, creating a friendly atmosphere. Also for kitchen decor or wall art for personal coffee stations at home,} \usecase{enhancing the cozy vibe} \\
\bottomrule
\end{tabularx}
\end{table*}

\begin{table*}[htbp]
\centering
\caption{Evaluation prompts (ID: 16-30). \fontstyle{Font styles} are indicated in \fontstyle{brown} and \usecase{use cases} in \usecase{blue}.}
\label{tab:prompts-2}
\small
\setlength{\tabcolsep}{6pt}
\begin{tabularx}{\textwidth}{@{}c >{\raggedright\arraybackslash}X@{}}
\toprule
\textbf{ID} & \textbf{Prompt} \\
\midrule
16 & \fontstyle{Handwritten, rough, textured and brush script.} \usecase{Used in social media graphics, promotional content, and blog headers for lifestyle, travel, or personal brands} \\
17 & \usecase{Suitable for posters or prints with motivational quotes for decor or gifts. Creative branding for businesses in arts, crafts, or entertainment} \\
18 & \usecase{Youth-oriented, for merchandise like t-shirts, stickers, and accessories for a younger audience} \\
19 & \fontstyle{Playful and vibrant, rounded font.} \usecase{Used for restaurant branding, pizza menus, and promotional materials} \\
20 & \fontstyle{Sans-serif font, modern, wide, sleek and geometric shape.} \usecase{Suitable for futuristic and tech-oriented design projects, as technology companies or innovative startups branding, and futuristic products} \\
21 & \fontstyle{Elegant, serif, modern.} \usecase{Logos for high-end clothing brands, upscale boutiques, elegant fashion lines. Books for lifestyle or beauty brands, product packaging} \\
22 & \usecase{Horror-themed and dramatic design projects, Halloween decorations signage, invitations, or party materials} \\
23 & \fontstyle{Beautiful display font, decorative,} \usecase{Suitable for academic presentation, friendly, smart, geometric} \\
24 & \fontstyle{Sans, creative, branding, bold, ligature, logo font, vintage, display font} \\
25 & \fontstyle{Signature, handwritten, ligature, unique, casual script, brush, beautiful, romantic} \\
26 & \fontstyle{Futuristic, neon style font. Typography designed with beautiful out-glow effect} \\
27 & \fontstyle{Dreamy, fancy, strange and surreal font. Typography designed with distorted effect} \\
28 & \fontstyle{Stroke, wild, handwritten, Chinese brush style.} \usecase{For antique-style decorative painting} \\
29 & \fontstyle{A childlike, rough, and untidy chalkboard style.} \usecase{Perfect for children's books, posters, casual projects, and any design needing a fun, vintage chalk look} \\
30 & \fontstyle{A bold, all-caps font,} \usecase{perfect for propaganda posters or modern vintage graphic designs. It comes in two styles: simple and grunge} \\
\bottomrule
\end{tabularx}
\end{table*}
 \fi

\end{document}